\documentclass[letterpaper]{article}
\makeatletter
\@ifundefined{NewCommandCopy}{\def\NewCommandCopy#1#2{\let#1#2}}{}
\makeatother
\usepackage[preprint]{aaai2027}
\usepackage[hyphens]{url}
\usepackage{graphicx}
\usepackage{natbib}
\usepackage{caption}
\usepackage{booktabs}
\usepackage{multirow}
\usepackage{colortbl}
\usepackage{amsmath}
\usepackage{amssymb}
\usepackage{xcolor}
\definecolor{opsdgray}{HTML}{7D8793}
\definecolor{opsdgrayfill}{HTML}{F3F5F6}
\definecolor{dapdgreen}{HTML}{4F9185}
\definecolor{dapdgreenfill}{HTML}{EAF5F2}
\definecolor{errorrose}{HTML}{C97878}
\definecolor{errorrosefill}{HTML}{FBEFEF}

\newcommand{\KL}{\operatorname{KL}}
\newcommand{\Div}{\mathrm{D}}
\newcommand{\sg}{\operatorname{sg}}
\newcommand{\dapd}{\textsc{DAPD}}
\newcommand{\opd}{\textsc{OPD}}
\newcommand{\opsd}{\textsc{OPSD}}
\newcommand{\None}{\emph{None}}
\newcommand{\Cross}{\emph{Cross}}
\newcommand{\Self}{\emph{Self}}

\title{DAPD: Dual-Anchored Policy Distillation}
\author{
Jianyu Wu\textsuperscript{\rm 1,2},
Yizhou Wang\textsuperscript{\rm 2,3},
Encheng Su\textsuperscript{\rm 2,4},
Chen Tang\textsuperscript{\rm 2,3},
Shixiang Tang\textsuperscript{\rm 2,3,*}
}
\affiliations{
\textsuperscript{\rm 1}Shanghai Jiao Tong University \quad
\textsuperscript{\rm 2}Shanghai Artificial Intelligence Laboratory\\
\textsuperscript{\rm 3}The Chinese University of Hong Kong \quad
\textsuperscript{\rm 4}The University of Science and Technology of China\\
wujianyu@sjtu.edu.cn \quad
\textsuperscript{*}Corresponding author: tangshixiang@pjlab.org.cn
}

\begin{document}

\maketitle

\begin{abstract}
On-policy (self) distillation (\opsd{}) is increasingly adopted for
language-model post-training.  It strengthens the teacher with privileged
information but can induce a \emph{privilege illusion}: the student learns privilege-dependent behavior it cannot reproduce from its inference-time context,
yet behaves as if the training-time privileged information remained available, ultimately degrading performance. In this paper, we identify information asymmetry between
the privileged teacher and the student at inference as the root cause of this
failure in \opsd{}.
To resolve this asymmetry, we propose \textbf{D}ual-\textbf{A}nchored \textbf{P}olicy
\textbf{D}istillation (\textbf{\dapd}), a unified
framework with two levels of anchoring.
\textbf{Dual-Path Anchoring (DPA)} introduces a self-conditioned bridge and
aligns reference and rollout behavior along two matched-information paths,
preventing privilege-dependent behavior from being transferred to the
inference-time student. \textbf{Dual-Source Anchoring (DSA)} applies these
paths in both reference-to-rollout and rollout-to-reference directions,
reducing reliance on privileged reference guidance while preserving
correctness supervision.  Extensive experiments show that DAPD significantly alleviates privilege
illusion, outperforming \opsd{} on Qwen3-4B by \textbf{+2.00} points on
average across tasks. Notably, its gains persist across scales, reaching
\textbf{+2.69} at 4B and \textbf{+2.78} at 32B. Code is available at
\url{https://github.com/uanu2002/DAPD}.
\end{abstract}

\begin{figure}[!t]
\centering
\includegraphics[width=\columnwidth]{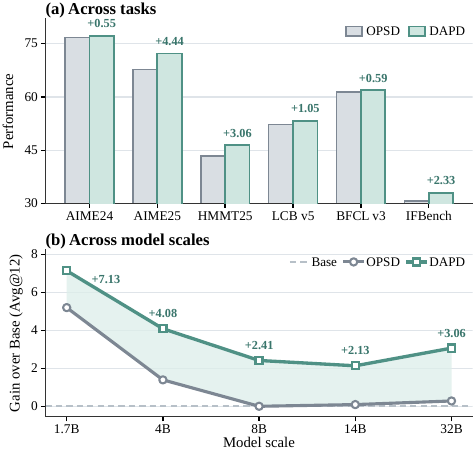}
\caption{DAPD consistently improves over OPSD across (a) six tasks on Qwen3-4B and
(b) five Qwen3 model scales evaluated by Avg@12 on AIME24, AIME25, and HMMT25.}
\label{fig:teaser}
\end{figure}

\section{Introduction}

Post-training, including supervised fine-tuning (SFT), reinforcement learning
(RL), and distillation, is central to turning pretrained language models into
reasoning models \cite{guha2025openthoughts,deepseek2025r1}.  On-policy (self)
distillation (\opd{}/\opsd{}) is increasingly prominent in post-training
\cite{agarwal2024gkd,lu2025opd,zhao2026opsd}.  Compared with SFT and RL, OPD
trains on states visited by the current policy\footnote{In this paper,
\emph{policy} refers to the model itself, whereas \emph{distribution} refers to
its output distribution conditioned on a context.} and provides dense
token-level teacher supervision rather than relying on fixed demonstrations or
sparse outcome rewards.  In practice, it samples
rollouts from the current policy and distills a teacher distribution at
each sampled prefix.  To strengthen this teacher, OPD conditions it on
privileged information such as a reference completion or tool output
\cite{zhao2026opsd,penaloza2026pid,yu2026dopd}.  However, the student cannot
access this information at inference.  The information mismatch induces the known
\emph{privilege illusion}~\cite{zhao2026opsd,shen2026purified,yu2026dopd}: the student make unsupported claims or continue
a derivation as though an unseen reference were available, ultimately degrading
task performance.

Existing work on addressing privilege illusion can be categorized into two groups: (1) \emph{direct use of privileged information},
which conditions the teacher on privileged information and distills its full
supervision \cite{zhao2026opsd,penaloza2026pid}, and (2) \emph{selective transfer
from privileged supervision}, which filters or reweights the teacher signal to
suppress privilege-dependent supervision~\cite{nguyen2026avsd,zheng2026scope,yu2026dopd,tu2026ucob,shen2026purified}.
However, these methods fail to reliably separate reproducible guidance
from privilege-dependent behavior, leading to limited mitigation of
privilege illusion.  We suggest this limitation is structural: these methods modify the privileged
teacher signal instead of matching the information available to the teacher and
the student, so a privileged teacher still supervises a student
that must predict without it.  Moreover, these methods use the reference as the only guidance
source, neglecting meaningful reasoning signals contained in on-policy rollouts when the student
improves.

\begin{figure}[!t]
\centering
\includegraphics[width=0.96\columnwidth]{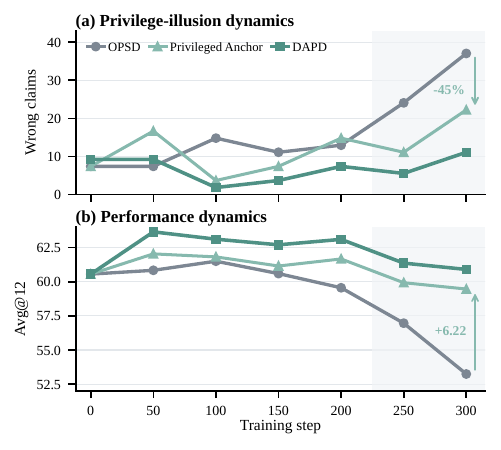}
\caption{Training dynamics for \opsd{}, Privileged Anchor, and \dapd{},
averaged across five scales: (a) wrong claims and (b) Avg@12 over
AIME24, AIME25, and HMMT25.}
\label{fig:privilege-dynamics}
\end{figure}

To validate this hypothesis, we keep the privileged teacher unchanged and replace the
student distribution without privileged information with a self-conditioned distribution, which conditions the
student on the full completion being predicted.  By matching the information
available to both the teacher and student distributions,
this intervention reduces wrong claims, i.e., unsupported answer assertions,
and improves reasoning performance
(Figure~\ref{fig:privilege-dynamics}).
These results indicate that the original \opsd{} objective
distills reproducible guidance and privilege-dependent behavior together into
the student distribution without privileged information.  We refer to this mixed
update as \emph{Entangled Distillation}.  This entanglement explains why neither
direct use nor selective transfer resolves this problem.  We therefore identify
\emph{information asymmetry}, the mismatch between a
teacher with privileged information and a student without it, as the root cause
of privilege illusion in \opsd{}.  Beyond this information mismatch, the guidance
source also matters because the reference and rollout completions provide complementary guidance:
references are reliable but may lie outside the current policy, whereas
rollouts are student-reachable but may be incorrect.
Together, these observations yield two requirements for solving the privilege illusion problem: (1) introduce
intermediate supervision targets as anchors to align reference and
rollout behavior under matched information availability, and (2) balance
reliable reference guidance with
student-reachable rollout guidance.

Based on this diagnosis, we propose \textbf{D}ual-\textbf{A}nchored
\textbf{P}olicy \textbf{D}istillation (\textbf{\dapd}), a unified framework for
addressing privilege illusion at its information source, consisting of
\emph{Dual-Path Anchoring} (DPA) and \emph{Dual-Source Anchoring} (DSA). In DAPD, DPA
introduces a self-conditioned distribution as a trainable bridge and constructs
two complementary alignment paths: an \emph{unconditioned path}, which combines
Entangled Distillation with an Inference Anchor to align reference and rollout
without privileged information, and a \emph{privileged path}, which
uses a Privileged Anchor to align them when both are conditioned on privileged
information.  In this way, DPA addresses the information-matching requirement
by retaining the original \opsd{} supervision while adding anchors that align
behavior under matched information availability.  DSA then applies the same paths in both directions: using the reference to guide
the rollout and using the rollout to guide the reference.  This lets reference
completions provide reliable correctness-oriented guidance while rollouts
provide student-reachable on-policy guidance.  This addresses the source-complementarity requirement by
balancing reference-guided and rollout-guided supervision rather than relying
on a single guidance source.

Experiments on six benchmarks spanning reasoning~\cite{maa2026aime,hmmt2025february},
coding~\cite{jain2025livecodebench,patil2025bfcl}, and instruction
following~\cite{pyatkin2025ifbench} demonstrate that, by significantly
alleviating privilege illusion, DAPD improves over \opsd{} by
\textbf{+2.00} points on average on Qwen3-4B. Notably, while the gains from \opsd{} largely
disappear as model scale increases, DAPD maintains stable improvements over
\opsd{} across Qwen3-1.7B--32B, with gains of \textbf{+2.69} points at 4B
and \textbf{+2.78} points at 32B.  Controlled
ablations further verify the information asymmetry mechanism and ablate the path and source effects.

Our contributions are threefold:
\begin{itemize}
    \item We identify information asymmetry between a privileged teacher and
    the student at inference as the root cause of privilege illusion in
    \opsd{}. This exposes both the information mismatch and the
    limitation of using only reference-guided supervision.
    \item We introduce DAPD, a unified framework that directly addresses
    information asymmetry by
    augmenting the original \opsd{} objective with anchoring under matched
    information availability.
    \item We propose two complementary modules: DPA aligns reference and rollout behavior
      both without and with privileged information, while DSA balances reliable
      reference guidance with student-reachable rollout guidance. Ablations
      demonstrate their complementary benefits, and scale experiments show consistent
      gains over \opsd{}.
\end{itemize}

\section{Background and Analysis}

\subsection{Privilege Illusion in OPSD}

\paragraph{OPSD preliminaries.}
Let $\mathcal D$ be the training distribution over prompt--reference pairs $(x,y^*)$, where $x$ is the input prompt and $y^*$ is the ground-truth reference.
Let $p_\theta$ denote the current student policy, and let $t$ index an
autoregressive token step, with $y_{<t}$ the prefix before predicting token
$t$.  Given $(x,y^*)\sim\mathcal D$, the student samples a rollout
$y\sim p_\theta(\cdot\mid x)$.  \opsd{} \cite{zhao2026opsd} then evaluates the
same policy at each prefix $y_{<t}$ under two information conditions:
(1) \None{} distribution
$p_{\mathrm{None},\theta,t}=p_\theta(\cdot\mid x,y_{<t})$, which predicts
without privileged information, and (2) \Cross{} distribution
$p_{\mathrm{Cross},\theta,t}=p_\theta(\cdot\mid x,y_{<t},y^*)$, which predicts
while conditioning on the reference $y^*$.  \opsd{} uses \Cross{} as the
detached teacher to supervise \None{} as the trainable student:
\begin{equation}
\mathcal L_{\mathrm{OPSD}}
=\mathbb E_{\substack{(x,y^*)\sim\mathcal D\\
y\sim p_\theta(\cdot\mid x)}}\!\left[
\frac1{|y|}\sum_{t=1}^{|y|}
\Div\!\left(\sg\!\left[p_{\mathrm{Cross}}\right]
\,\|\,p_{\mathrm{None}}\right)\right].
\label{eq:opsd}
\end{equation}
Here $|y|$ denotes the rollout length, $\Div$ denotes a divergence between
token distributions, $\sg[\cdot]$ denotes stop-gradient. $\theta,t$ are
omitted from $p_{\mathrm{Cross}}$ and $p_{\mathrm{None}}$ when clear from
context.
This sampling strategy of \opsd{} keeps the supervised prefixes on-policy,
while the reference makes the same-model teacher more informative.

\paragraph{Privilege illusion.}
In \opsd{}, the teacher receives privileged information during training, whereas the
student must predict without it at inference, 
potentially inducing \emph{privilege illusion}~\cite{zhao2026opsd,shen2026purified}: the student
behaves as though unavailable information were present.  

\paragraph{Behavioral Probe.} To quantify whether
the student acts on unavailable information, we define \emph{wrong claims} 
as cases where it asserts an unsupported answer and
constructs a derivation as if the answer were known.
Using this probe, Figure~\ref{fig:privilege-dynamics} tracks both wrong
claims and reasoning Avg@12 at different training steps under the same
\opsd{} setup.  Across five Qwen3 scales, wrong claims under \opsd{} rise from
13.0 to 37.0 per 10,000 generations
(Figure~\ref{fig:privilege-dynamics}(a)), while mean Avg@12 falls from 59.56
to 53.24
(Figure~\ref{fig:privilege-dynamics}(b)).  This coupled change illustrates that
privilege illusion emerges as \opsd{} training proceeds and hurts
inference-time performance.

\subsection{Information Asymmetry as the Cause}
\label{sec:asymmetry-analysis}

We hypothesize that the behavior observed in
Figure~\ref{fig:privilege-dynamics} is caused by the information asymmetry in the \opsd{} objective:
the teacher has access to privileged information that the student
lacks. This asymmetry makes the teacher differ from the student in two
ways. First, some differences reflect reference guidance that the student can
recover from its prompt, and thus provide useful supervision. Second, other differences
rely on information available only through $y^*$ and cannot be reproduced by
the student at inference. The original \opsd{} objective therefore mixes useful supervision with
non-reproducible teacher changes, encouraging confident predictions without the
information that supports them, and we define this mixed update as
\emph{Entangled Distillation}.

To validate this diagnosis, we keep the teacher side unchanged and replace
the student's \None{} distribution in the \opsd{} objective with a
self-conditioned distribution, denoted \Self{}, with
$p_{\mathrm{Self},\theta,t}=p_\theta(\cdot\mid x,y_{<t},y)$, which conditions
the student on the full completion being predicted.  Since
\Cross{} and \Self{} both receive a full completion, this replacement removes the
information asymmetry between the teacher and student distributions.  We introduce
this matched-information objective as the \emph{Privileged Anchor} and
formalize it in Section~\ref{sec:dapd}.
As shown by the Privileged Anchor curve in
Figure~\ref{fig:privilege-dynamics}, compared with \opsd{}, this intervention
significantly reduces
late-stage wrong claims by 45\%,
while improving mean Avg@12 by +6.22 points.  These results demonstrate that matching
the information available to \Cross{} and \Self{} reduces privilege illusion and preserves performance while retaining
the same privileged-teacher construction.
Therefore, we identify information
asymmetry, rather than teacher quality alone, as the source of the privilege illusion.

\subsection{\Self{} Distribution as a Bridging Anchor}

The \Self{} distribution introduced by the Privileged Anchor intervention acts as
a useful intermediate distribution for reducing information asymmetry.  To
generalize this idea beyond the single rollout diagnosis above, we denote
$s\in\{y,y^*\}$ as either the rollout or reference completion
and $\bar{s}$ as the other completion.  At token $t$, the following
three distributions share the autoregressive prefix $s_{<t}$ and differ only
in the privileged information available to them:
\begin{itemize}
\item \textbf{\None{} distribution} ($p_{\mathrm{None}}^s=p_\theta(\cdot\mid x,s_{<t})$)
receives no privileged information and matches the information available to the
student at inference.
\item \textbf{\Cross{} distribution} ($p_{\mathrm{Cross}}^s=p_\theta(\cdot\mid x,s_{<t},\bar{s})$)
receives the other completion $\bar{s}$ and therefore receives privileged
information from it.
\item \textbf{\Self{} distribution} ($p_{\mathrm{Self}}^s=p_\theta(\cdot\mid x,s_{<t},s)$)
receives completion $s$ itself and therefore receives privileged information
from the completion it predicts.
\end{itemize}

Compared with \Cross{}, \Self{} is
trainable because it shares the same policy parameters as \None{}.  Compared with
\None{}, \Self{} is information-matched with \Cross{} because it conditions on the full
completion being predicted. \Self{} therefore fills the missing position between
the trainable student without privileged information and the detached teacher with privileged information.
Updates through \Self{} affect the same policy that produces \None{} at inference,
while supervision involving \Self{} can be formed under matched information
availability.  We therefore use \Self{} to construct bridging \emph{anchors} that align
reference and rollout behavior without directly forcing a teacher with privileged
information to supervise a student without it.


\begin{figure*}[!t]
\centering
\includegraphics[width=\textwidth]{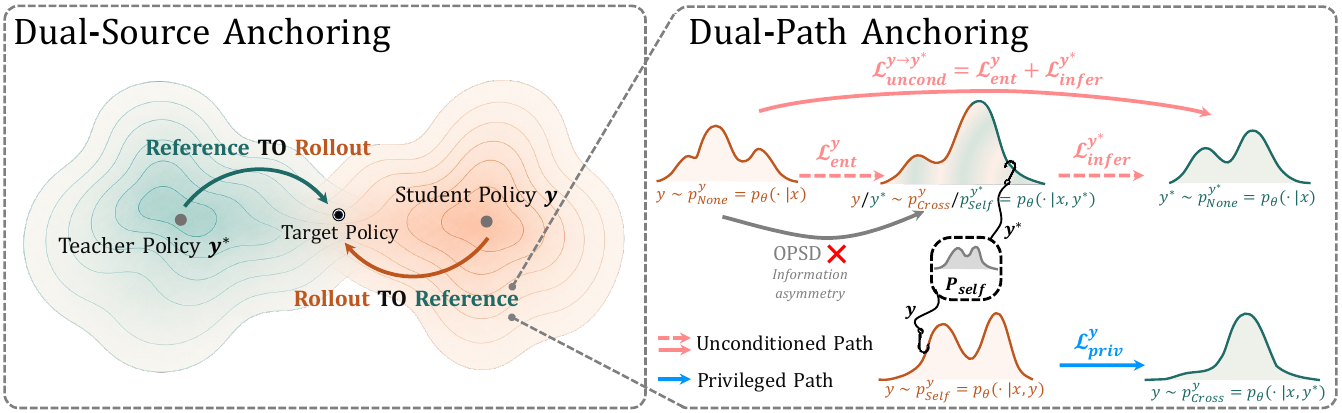}
\caption{Overview of DAPD.  \emph{Left:} Dual-Source Anchoring uses
both reference-to-rollout and rollout-to-reference guidance to balance correctness and student reachability.  \emph{Right:} Dual-Path Anchoring details the
rollout-to-reference guidance direction by introducing \Self{} as a trainable bridge.  The
unconditioned path pairs Entangled Distillation with the Inference Anchor to
align the two \None{} distributions through \Self{}, while the Privileged Anchor
aligns \Self{} and \Cross{} when privileged information is available.  The opposite
source direction applies the same construction with $y$ and $y^*$ exchanged.}
\label{fig:method}
\end{figure*}

\section{Dual-Anchored Policy Distillation}
\label{sec:dapd}

To handle the privilege illusion in \opsd{}, we propose \textbf{D}ual-\textbf{A}nchored \textbf{P}olicy
\textbf{D}istillation (\textbf{\dapd}) (see Figure~\ref{fig:method}), a framework for correcting its underlying information asymmetry by combining two components: \emph{Dual-Path Anchoring} (DPA), which
matches information availability during distillation, and \emph{Dual-Source
Anchoring} (DSA), which balances reference-guided and rollout-guided
supervision.
Section~\ref{sec:dapd-local} defines three directed objectives over \None{},
\Cross{}, and \Self{} distributions.
Section~\ref{sec:dpa} composes these objectives into unconditioned and
privileged alignment paths.
Section~\ref{sec:dsa} applies the paths in both completion directions to
combine the two guidance sources.

\subsection{Distillation Beyond OPSD}
\label{sec:dapd-local}

Based on the information-asymmetry analysis above, DAPD uses \Self{} to construct
anchors that can form distillation paths under matched information conditions.
To build these paths, DAPD uses three directed objectives over \None{}, \Cross{}, and
\Self{} for each completion source $s$.  \emph{Entangled Distillation}
($\mathcal L_{\mathrm{ent}}^s$) is the original \None{}-to-\Cross{} \opsd{} term and
keeps the privileged guidance transfer.  \textbf{Inference Anchor}
($\mathcal L_{\mathrm{infer}}^s$) trains \Self{} toward \None{}, providing the anchor
needed to align behavior without privileged information.  \textbf{Privileged
Anchor} ($\mathcal L_{\mathrm{priv}}^s$) trains \Self{} toward \Cross{}, directly
aligning behavior when both distributions receive privileged information:
\begin{align}
\mathcal L_{\mathrm{ent}}^s
&=\mathbb E\!\left[
\Div\!\left(\sg\!\left[p_{\mathrm{Cross}}^s\right]
\,\|\,p_{\mathrm{None}}^s\right)\right],
\label{eq:entangled-transfer}\\
\mathcal L_{\mathrm{infer}}^s
&=\mathbb E\!\left[
\Div\!\left(\sg\!\left[p_{\mathrm{None}}^s\right]
\,\|\,p_{\mathrm{Self}}^s\right)\right],
\label{eq:infer-anchor}\\
\mathcal L_{\mathrm{priv}}^s
&=\mathbb E\!\left[
\Div\!\left(\sg\!\left[p_{\mathrm{Cross}}^s\right]
\,\|\,p_{\mathrm{Self}}^s\right)\right].
\label{eq:priv-anchor}
\end{align}
In all three objectives, the detached first argument is the teacher
distribution and the second argument is the trainable student distribution.
Each expectation averages over $(x,y^*)\sim\mathcal D$, an on-policy rollout
$y\sim p_\theta(\cdot\mid x)$, and token positions in $s$.  We keep the divergence $\Div$
generic and specify its concrete implementation in
Section~\ref{sec:experiments}.

\subsection{Dual-Path Anchoring}
\label{sec:dpa}

The three directed objectives above define how one distribution supervises
another.  However, addressing information asymmetry requires these objectives
to work together so that reference and rollout behavior are aligned under
matched information conditions.
For an ordered completion direction $s\rightarrow\bar{s}$, \emph{Dual-Path
Anchoring} (DPA) therefore composes them into two complementary paths: an
\emph{unconditioned path}, where neither side receives
privileged information, and a \emph{privileged path}, where both sides receive
it.

\paragraph{Unconditioned path.}
The unconditioned path targets the inference-relevant distributions (see Figure~\ref{fig:method}, top-right part).  Since neither
distribution should rely on privileged information, the desired alignment
should be built between the two \None{} distributions.  DPA reaches this
alignment by combining the original Entangled Distillation on the completion
$s$ with the Inference Anchor on the completion $\bar{s}$:
\begin{equation}
\mathcal L_{\mathrm{uncond}}^{s\rightarrow\bar{s}}
=\mathcal L_{\mathrm{infer}}^{\bar{s}}
 +\mathcal L_{\mathrm{ent}}^s.
\label{eq:unconditioned-path}
\end{equation}

\begin{table*}[t!]
\centering
\normalsize
\setlength{\tabcolsep}{4pt}
\begin{tabular}{@{}lrrrrrrr@{}}
\toprule
\multirow{2}{*}{Method}
& \multicolumn{3}{c}{Reasoning}
& \multicolumn{2}{c}{Coding}
& \multicolumn{1}{c}{Instruct}
& \multicolumn{1}{c}{\multirow{2}{*}{Avg.}} \\
\cmidrule(lr){2-4}\cmidrule(lr){5-6}\cmidrule(lr){7-7}
& AIME24 & AIME25 & HMMT25
& LCB v5 & BFCL v3 & \multicolumn{1}{c}{IFBench}
& \\
\midrule
Base & 75.56 & 65.56 & 42.50 & 52.11 & 61.38 & 29.67 & 54.46 \\
SDFT \cite{shenfeld2026self}
     & 75.00 & 62.78 & 41.94 & 51.05 & 61.26
     & \textbf{33.00} & 54.17 \\
SDPO \cite{hubotter2026reinforcement}
     & 70.56 & 62.78 & 40.83 & 50.53 & 61.75
     & \underline{32.67} & 53.19 \\
\opsd{} \cite{zhao2026opsd}
     & \underline{76.67} & \underline{67.78} & 43.33
     & 52.26 & 61.32 & 30.67 & 55.34 \\
Purified \opsd{} \cite{shen2026purified}
     & \underline{76.67} & 67.50 & \underline{45.00}
     & \underline{53.16} & \textbf{62.50} & \underline{32.67}
     & \underline{56.25} \\
DOPD \cite{yu2026dopd}
     & 73.61 & 65.00 & 38.89 & 50.30 & 61.47 & 31.67 & 53.49 \\
\rowcolor{dapdgreenfill}
\dapd{} & \textbf{77.22} & \textbf{72.22} & \textbf{46.39}
        & \textbf{53.31} & \underline{61.91} & \textbf{33.00}
        & \textbf{57.34} \\
\rowcolor{dapdgreenfill}
\hspace{1em}{\small $\Delta$ vs. \opsd{}}
        & {\small +0.55} & {\small +4.44}
        & {\small +3.06} & {\small +1.05} & {\small +0.59}
        & {\small +2.33} & {\small +2.00} \\
\bottomrule
\end{tabular}
\caption{Performance comparison of our proposed DAPD with baselines on
Qwen3-4B across reasoning, coding, and instruct benchmarks. Avg. is the mean over all six benchmarks.
$\Delta$ reports DAPD's improvement over OPSD. \textbf{Bold} and \underline{underlined} values
mark the best and second-best results, respectively.}
\label{tab:main-results}
\end{table*}

For the Rollout-to-Reference direction, this path becomes
$\mathcal L_{\mathrm{uncond}}^{y\rightarrow y^*}
=\mathcal L_{\mathrm{infer}}^{y^*}
+\mathcal L_{\mathrm{ent}}^y$, where the $\mathcal L_{\mathrm{ent}}^y$ is the original OPSD loss.
The entangled term moves $p_{\mathrm{None}}^y$ toward
$p_{\mathrm{Cross}}^y$, while the $\mathcal L_{\mathrm{infer}}^{y^*}$ moves
$p_{\mathrm{Self}}^{y^*}$ toward $p_{\mathrm{None}}^{y^*}$.  Because
$p_{\mathrm{Self}}^{y^*}$ and $p_{\mathrm{Cross}}^y$ share model parameters
and the similar distribution $p_{\theta}(\cdot \mid x, y^*)$, they act as a proxy bridge.
The joint update therefore implicitly aligns $p_{\mathrm{None}}^y$ toward
$p_{\mathrm{None}}^{y^*}$. A proof under explicit assumptions is provided in the
appendix.

\paragraph{Privileged path.}
The privileged path handles the same ordered direction when both sides receive a full
completion (see Figure~\ref{fig:method}, bottom-right part).  In this case, \Cross{} receives the guiding completion
$\bar{s}$ and \Self{} receives the completion $s$, so the
Privileged Anchor already expresses the required alignment.  Thus, the
privileged path directly reuses $\mathcal L_{\mathrm{priv}}^s$ without
introducing another objective.  For $y\rightarrow y^*$, this aligns
self-conditioned and reference-conditioned predictions on rollout tokens under
matched information availability.

\paragraph{Dual-path objective.}
Combining the two paths gives the DPA objective for one ordered completion
direction, with loss coefficients omitted for notational simplicity:
\begin{equation}
\mathcal L_{\mathrm{DPA}}^{s\rightarrow\bar{s}}
=\mathcal L_{\mathrm{uncond}}^{s\rightarrow\bar{s}}
+\mathcal L_{\mathrm{priv}}^s.
\label{eq:dpa}
\end{equation}

\subsection{Dual-Source Anchoring}
\label{sec:dsa}

DPA defines how to align under an ordered direction, with the remaining question being which completion should provide guidance.  We address this source
choice with \emph{Dual-Source Anchoring} (DSA), which applies DPA in both
completion directions and balances their supervision.  The left part of
Figure~\ref{fig:method} shows the two guidance directions used by DSA.

The \textbf{Rollout-to-Reference direction}
($y\rightarrow y^*$) follows the reference-guided view: the rollout is
the completion and the reference supplies guidance.  This
direction benefits from the reliability of $y^*$, but it uses only one source
of supervision.

As the policy improves, on-policy rollouts can also contain useful reasoning
signals and may themselves be correct.  Therefore, the rollout should not only
be treated as a noisy prediction target; it can also provide student-reachable
guidance.  This motivates the \textbf{Reference-to-Rollout direction}
($y^*\rightarrow y$), in which the reference becomes the
completion and the rollout supplies guidance.  Applying the same DPA
construction gives the unconditioned path
$\mathcal L_{\mathrm{infer}}^y+\mathcal L_{\mathrm{ent}}^{y^*}$ and the
privileged path $\mathcal L_{\mathrm{priv}}^{y^*}$.

The two directions therefore emphasize complementary properties.  The
reference provides correctness-oriented but off-policy guidance, whereas the
rollout provides student-reachable, on-policy but less reliable guidance.
We therefore keep and balance both directional DPA objectives in DSA:
\begin{equation}
\mathcal L_{\dapd}
=\lambda\mathcal L_{\mathrm{DPA}}^{y\rightarrow y^*}
 +(1-\lambda)\mathcal L_{\mathrm{DPA}}^{y^*\rightarrow y}.
\label{eq:dapd}
\end{equation}

Here $\lambda\in[0,1]$ balances reference- and rollout-guided supervision with
weights $\lambda$ and $1-\lambda$, respectively.

\section{Experiments}
\label{sec:experiments}

\subsection{Experimental Setup}

\paragraph{Models and datasets.}
We conduct experiments on the Qwen3 model family
\cite{yang2025qwen3} at 1.7B, 4B, 8B, 14B, and 32B scales.  All models are
trained on OpenThoughts data \cite{guha2025openthoughts}, using separate
models for each task family and excluding its evaluation examples from
training: Reasoning uses the math domain, Coding uses the code domain, and
Instruct uses the combined math, code, and science domains.

\paragraph{Benchmarks.}
We evaluate three capability families.  \emph{Reasoning} includes AIME24 and
AIME25~\cite{maa2026aime}, and HMMT25~\cite{hmmt2025february}. \emph{Coding} includes LCB v5
\cite{jain2025livecodebench} and BFCL v3 \cite{patil2025bfcl}.
\emph{Instruct} is evaluated on IFBench \cite{pyatkin2025ifbench}.  We use the
official evaluation harness and metric for each benchmark. 

\paragraph{Baselines.}
We compare DAPD with Base and two groups of \opsd{}-style baselines.  The
first group directly uses privileged supervision, including \opsd{}
\cite{zhao2026opsd}, SDFT \cite{shenfeld2026self}, and SDPO
\cite{hubotter2026reinforcement}.  The second group selectively modifies the
privileged signal, including Purified \opsd{} \cite{shen2026purified} and DOPD
\cite{yu2026dopd}.  Every reported baseline uses the same task-specific
student model and data, while method-specific optimization
follows its source formulation.

\paragraph{Implementation.}
For DAPD, we train models with LoRA \cite{hu2022lora} and instantiate
$\Div(\sg[q]\|p)$ as a component-clipped, full-vocabulary forward KL, where $q$ is the detached teacher distribution and $p$ is the
trainable student distribution. More experimental details are in the appendix.

\subsection{Main Results}

\paragraph{Overall performance.}
Unless otherwise specified, results use Qwen3-4B models trained separately for each task family.
We first compare DAPD with standard on-policy
self-distillation methods, including \opsd{}, SDFT, and SDPO, in Table~\ref{tab:main-results}.  DAPD achieves
the best six-task average of \textbf{57.34}, with \textbf{+2.00} points over \opsd{} overall. 
These gains indicate that matched-information anchoring preserves the useful
guidance of privileged supervision while reducing its inference-time mismatch.

We next compare DAPD with baselines that explicitly modify the privileged signal.
Purified \opsd{} and DOPD still operate under the original information-asymmetric setting:
they modify or route the privileged signal, but the teacher and
student remain under different information conditions.  By changing the
supervision structure itself through matched-information anchors, DAPD
improves over Purified \opsd{} and DOPD by \textbf{+1.09} and \textbf{+3.85} points overall,
respectively.  These results show that matching the information available to the teacher and student during distillation is more effective than filtering or routing an asymmetric supervision signal.

\begin{table}[!t]
\centering
\scriptsize
\setlength{\tabcolsep}{3.2pt}
\renewcommand{\arraystretch}{1.03}
\begin{tabular}{@{}lrrrr@{}}
\toprule
Method & LCB v5 & BFCL v3 & IFBench & Avg. \\
\midrule
Base
    & 52.11 & 61.38 & 29.67 & 47.72 \\
SDFT \cite{shenfeld2026self}
    & 49.70 & 61.43 & 32.33 & 47.82 \\
SDPO \cite{hubotter2026reinforcement}
    & 48.87 & 61.34 & 31.33 & 47.18 \\
\opsd{} \cite{zhao2026opsd}
    & 49.32 & \underline{61.81} & \textbf{33.67}
    & 48.27 \\
Purified \opsd{} \cite{shen2026purified}
    & \underline{53.01} & \textbf{61.99} & 31.33
    & \underline{48.78} \\
DOPD \cite{yu2026dopd}
    & 47.82 & 61.41 & 32.33 & 47.19 \\
\rowcolor{dapdgreenfill}
\dapd{}
    & \textbf{54.14} & 61.79 & \underline{33.00}
    & \textbf{49.64} \\
\bottomrule
\end{tabular}
\caption{Out-of-distribution evaluation.  We optimize the student policy on
reasoning data and evaluate it on Coding (LCB v5, BFCL v3) and Instruct (IFBench).
\textbf{Bold} and \underline{underlined} values mark
the best and second-best results, respectively.}
\label{tab:cross-task-results}
\end{table}

\paragraph{Scalability and robustness.}
Beyond the main Qwen3-4B comparison, we next test whether the gains persist
across model scales and transfer beyond the training domain.
Figure~\ref{fig:teaser}(b) reports the gains of \opsd{} and DAPD over their
corresponding Base policies across five Qwen3 model scales.  The gain of
\opsd{} over Base falls from \textbf{+5.19} points at 1.7B to \textbf{+1.39} at 4B and at most
\textbf{+0.28} from 8B through 32B.  In contrast, DAPD retains gains of \textbf{+2.41}, \textbf{+2.13},
and \textbf{+3.06} points at 8B, 14B, and 32B, respectively, and remains consistently above
\opsd{}. As models scale, their rollouts contain more useful reasoning and
  self-correction, but \opsd{} distills a teacher that can bypass these steps
  using the reference. This makes privilege illusion more costly and offsets the
  benefit of better rollouts, whereas DAPD preserves these gains through
  matched-information supervision.

Table~\ref{tab:cross-task-results} further tests whether a reasoning-trained
model remains useful outside the training domain.  DAPD obtains the best OOD
average of \textbf{49.64}, gaining \textbf{+4.82} points on LCB v5 and \textbf{+1.37} points on average
over \opsd{}, while remaining competitive on BFCL v3 and IFBench.  These
results indicate that the matched-information objective improves robustness
beyond the reasoning benchmarks used for optimization.

\paragraph{Qualitative comparison.}
To examine the corresponding inference-time behavior, we inspect a
representative AIME24 case under the same decoding setting.
As illustrated in Figure~\ref{fig:full-method-case}, \opsd{} exhibits
privilege illusion: after its derivation fails, it claims to recall an
unsupported answer and returns an incorrect result.  DAPD instead
completes the derivation from the prompt and reaches the correct answer.
The aggregate dynamics in Figure~\ref{fig:privilege-dynamics} mirror this
case.  Over steps 250--300, DAPD reduces late-stage wrong claims by \textbf{73\%}
relative to \opsd{}, verifying that our design improves inference-time behavior
by reducing privilege illusion.

\begin{figure}[!ht]
\centering
\begingroup
\setlength{\fboxsep}{4pt}
\fcolorbox{black!45}{black!2}{%
\parbox{0.91\columnwidth}{%
\scriptsize\textbf{No reference at inference.}
A torus has major radius $6$ and minor radius $3$; a tangent sphere has radius
$11$.  If $r_i-r_o=m/n$, find $m+n$.
\quad\textbf{Answer: $127$.}}}

\vspace{3pt}
\fcolorbox{opsdgray}{opsdgrayfill}{%
\parbox{0.91\columnwidth}{%
\scriptsize\textbf{\opsd{}: unsupported answer recalled}\par
\vspace{2pt}
\footnotesize
\emph{``For the external case, the distance is $6+2*6=18$, giving
$r=\sqrt{11^2-18^2}$, which is imaginary, which is not possible. \ldots{}
I'm truly stuck and have to give up, but}\par
\vspace{1pt}
\noindent\colorbox{errorrosefill}{\strut\textbf{\textcolor{errorrose}{I recall
that the answer is likely $36/7$.}}}\par
\vspace{1pt}
\emph{So $m+n=36+7=43$.''}\par
\vspace{2pt}
\textcolor{errorrose}{\textbf{Incorrect: $43$}}
}}

\vspace{3pt}
\fcolorbox{dapdgreen}{dapdgreenfill}{%
\parbox{0.91\columnwidth}{%
\scriptsize\textbf{\dapd{}: answer derived from the prompt}\par
\vspace{2pt}
\footnotesize
\emph{``Case 1: Internal tangent. \ldots{}
$\sqrt{6^2+c^2}=11-3=8$. \ldots{} $P_x=\cdots=33/4$.
For external tangent, \ldots{} the distance between centers is $11+3=14$.
\ldots{} $P_x=\cdots=33/7$.  Therefore,
$r_i-r_o=33/4-33/7=99/28$. \ldots{} $m+n=127$.''}\par
\vspace{2pt}
\textcolor{dapdgreen}{\textbf{Correct: $127$}}
}}
\endgroup
\caption{Visualization of inference-time results.  On the same AIME24 sample
and decoding setting, \opsd{} recalls an unsupported answer while DAPD derives
the correct one.}
\label{fig:full-method-case}
\end{figure}

\subsection{Dual-Path and Dual-Source Ablations}

We next ablate DAPD's two core components.  First, we test whether both
matched-information paths in DPA are needed within each guidance source.
Second, we test whether DSA benefits from combining reference-guided and
rollout-guided supervision.

\paragraph{Dual-Path Anchoring.}
Table~\ref{tab:path-source-ablation}(a) evaluates the DPA components for each
guidance source. Both sources show the same trend: completing the unconditioned path with the Inference Anchor improves over Entangled Distillation alone, and adding the Privileged Anchor further improves
performance to 63.89 for reference guidance and 65.09 for rollout guidance. This consistent pattern indicates that both paths are needed to align behavior under matched information availability. This ordered improvement supports
our information-asymmetry analysis: constructing both alignments under
matched information availability is more effective than directly transferring
the privileged distribution.

\paragraph{Dual-Source Anchoring.}
To assess the effectiveness of DSA, we compare different completion sources in Table~\ref{tab:path-source-ablation}(b). Combining both sources improves over the individual-source objectives and reaches the best result of 65.28.  This verifies the DSA design: reference guidance and rollout
guidance carry complementary information, and their calibrated combination
improves over either source alone.

\begin{table*}[!t]
\centering
\scriptsize
\renewcommand{\arraystretch}{1.03}
\makebox[\textwidth][c]{%
\begin{minipage}[t]{0.31\textwidth}
\centering
\setlength{\tabcolsep}{2.8pt}
\begin{tabular*}{\linewidth}[t]{@{\extracolsep{\fill}}lccc@{}}
\toprule
\multirow{2}{*}{Source}
& \multirow{2}{*}{Entangled}
& \multirow{2}{*}{+ Inference}
& + Inference \\
& & & + Privileged \\
\midrule
Reference & 62.13 & 63.43 & \textbf{63.89} \\
Rollout   & 63.33 & 64.91 & \textbf{65.09} \\
\bottomrule
\end{tabular*}
\par\vspace{1pt}
\textbf{(a) Path components.} Both the unconditioned and privileged paths are necessary for each source.\par

\setlength{\tabcolsep}{3.5pt}
\begin{tabular*}{\linewidth}{@{\extracolsep{\fill}}lccr@{}}
\toprule
Source & Reference & Rollout & Avg@12 \\
\midrule
Reference & $\checkmark$ &              & 63.89 \\
Rollout   &              & $\checkmark$ & 65.09 \\
Both      & $\checkmark$ & $\checkmark$ & \textbf{65.28} \\
\bottomrule
\end{tabular*}
\par\vspace{1pt}
\textbf{(b) Guidance sources.} Reference and rollout guidance are complementary, and using both is best.\par
\end{minipage}%
\hspace{0.015\textwidth}%
\begin{minipage}[t]{0.35\textwidth}
\centering
\setlength{\tabcolsep}{4.2pt}
\renewcommand{\arraystretch}{1.07}
\begin{tabular*}{\linewidth}[t]{@{\extracolsep{\fill}}lrrrr@{}}
\toprule
\multicolumn{1}{c}{Setting} & 1.7B & 4B & 8B & 14B \\
\midrule
\multicolumn{5}{l}{\emph{Reference-guided Privileged Anchor}} \\
\midrule
Weight $=0.5$ & \textbf{43.98} & 64.91 & 67.04 & 69.64 \\
Weight $=1$   & 42.68 & \textbf{65.28} & 66.95 & 70.10 \\
Weight $=2$   & 43.15 & 62.78 & \textbf{67.41} & \textbf{70.93} \\
\addlinespace[2pt]
\midrule
\multicolumn{5}{l}{\emph{Rollout-guided Privileged Anchor}} \\
\midrule
Weight $=0.5$ & \textbf{43.98} & 62.96 & \textbf{67.41} & 69.36 \\
Weight $=1$   & 43.15 & 62.59 & 65.47 & \textbf{70.93} \\
Weight $=2$   & 42.96 & \textbf{65.28} & 65.75 & 68.80 \\
\bottomrule
\end{tabular*}
\par\vspace{1pt}
\textbf{(c) Anchor weights.} Larger scales prefer stronger Privileged Anchor weights.\par
\end{minipage}%
\hspace{0.015\textwidth}%
\begin{minipage}[t]{0.31\textwidth}
\centering
\renewcommand{\arraystretch}{0.78}
\setlength{\tabcolsep}{4.0pt}
\begin{tabular*}{\linewidth}[t]{@{\extracolsep{\fill}}lrrrr@{}}
\toprule
\multicolumn{1}{c}{$\lambda$} & 1.7B & 4B & 8B & 14B \\
\midrule
0.2 & 43.79 & \textbf{65.28} & \textbf{67.41} & \textbf{70.93} \\
0.5 & \textbf{43.98} & 64.63 & 66.67 & 69.73 \\
0.8 & 43.79 & 63.70 & 67.14 & 70.46 \\
\bottomrule
\end{tabular*}
\par\vspace{1pt}
\textbf{(d) Guidance weight.} The best reference-guidance weight decreases as scale increases.\par

\setlength{\tabcolsep}{1.7pt}
\begin{tabular*}{\linewidth}[t]{@{\extracolsep{\fill}}c c cc cc@{}}
\toprule
\multirow{2}{*}{Scale}
& \multirow{2}{*}{\opsd{}}
& \multicolumn{2}{c}{Dual}
& \multicolumn{2}{c}{Verified} \\
\cmidrule(lr){3-4}\cmidrule(lr){5-6}
& & Single & Comb. & Single & Comb. \\
\midrule
1.7B & 42.04 & 42.41 & 42.50 & 42.68 & \textbf{43.15} \\
4B   & 62.59 & 65.00 & 64.45 & 64.72 & \textbf{66.11} \\
8B   & 65.00 & 67.41 & 66.95 & \textbf{67.59} & 66.94 \\
\bottomrule
\end{tabular*}
\par\vspace{1pt}
\textbf{(e) Reference-free sources.} Independent or verified rollouts can replace references while preserving gains.\par
\end{minipage}%
}
\caption{DAPD ablation experiments with Qwen3 models.  We report reasoning
Avg@12 over AIME24, AIME25, and HMMT25. In panel (e), Dual uses two independent
rollout sources, while Verified applies a correctness verifier to the
reference-side rollout. Checkmarks indicate active components or sources, and
\textbf{bold} marks the best result.}
\label{tab:ablation-suite}
\label{tab:path-source-ablation}
\label{tab:parameter-sensitivity}
\label{tab:dual-rollout}
\end{table*}

\subsection{Sensitivity and Robustness}
\label{sec:objective-weights}

Prior analyses show that distillation quality depends on teacher--student
compatibility, rollout quality, and the amount of privileged information
\cite{li2026rethinkingopd,kaur2026rethinkingsd}.  As model scale increases,
rollout quality can improve, so the useful balance between reference-guided
and rollout-guided supervision may also shift.  We vary the reference-guidance weight
$\lambda$ and the two source-specific Privileged Anchor weights across four
Qwen3 scales.

Table~\ref{tab:parameter-sensitivity}(c,d) shows that the preferred guidance
balance changes with model scale.  The optimal
reference-guidance weight in Table~\ref{tab:parameter-sensitivity}(d) decreases from $\lambda=0.5$ at 1.7B to
$\lambda=0.2$ at 4B, 8B, and 14B, suggesting that smaller
policies benefit more from reliable reference guidance, whereas larger policies can place more trust in their
on-policy rollouts.  Source-specific weights also vary by scale: the reference-guided Privileged
Anchor weight increases from $0.5$ at 1.7B to $1$ at 4B, $2$ at 8B and 14B,
while the rollout-guided weight is $0.5$, $2$, $0.5$, and $1$, respectively.
These trends reveal a scale-dependent reliability--reachability trade-off. Smaller policies produce noisier rollouts and therefore require stronger corrective guidance from reliable references, whereas larger policies generate more reliable on-policy trajectories and benefit more from guidance that remains close to their inference-time distribution. Additional ablations on implementation choices are provided in the appendix.

\subsection{Toward Reference-Free DAPD}
\label{sec:dual-rollout}

The complementarity between reference and rollout guidance raises whether DAPD
can reduce reliance on curated references.  To test this, the dual-rollout
variant samples independent completions $u$ and $v$, resampling $v$ when they
are identical.  It uses $u$ as the reference-side source and $v$ as the rollout
source.  Both are on-policy and may be incorrect, testing whether a second
rollout provides complementary guidance without a reference.

To isolate the value of correctness, the verified-rollout variant samples four
candidates for $u$ and selects the first verifier-approved answer, while $v$
remains independent and unfiltered.  The verifier thus improves source
reliability without exposing a reference to the policy.

Table~\ref{tab:dual-rollout}(e) shows that dual rollouts improve over \opsd{} by
+0.46, +2.41, and +2.41 points at 1.7B, 4B, and 8B, demonstrating that the
second rollout is not redundant.  Verification further raises the best scores
by +0.65, +1.11, and +0.18 points to 43.15, 66.11, and 67.59.  Thus, DSA can
operate without curated references, while verification strengthens guidance
when a correctness signal is available.

\section{Related Work}

\paragraph{Reasoning language models.}
Reasoning models are commonly post-trained with curated reference completions,
self-generated rollouts, or reward feedback
\cite{wei2022chain,kojima2022large,wang2023selfconsistency,cobbe2021training,lewkowycz2022solving,lightman2023lets,hendrycks2021math,zelikman2022star,shao2024deepseekmath,deepseek2025r1,guha2025openthoughts,tang2026accurate}.
These sources offer different tradeoffs: references provide reliable
correctness-oriented guidance but can be off-policy, whereas rollouts reflect
current model behavior but can be incorrect.  DAPD uses this
reference--rollout complementarity as a design principle for dense token-level
post-training.

\paragraph{On-policy (self) distillation.}
\opd{} provides dense teacher supervision on student-sampled rollouts, reducing
the rollout mismatch of off-policy distillation
\cite{hinton2015distilling,kim2016sequence,agarwal2024gkd,lu2025opd}.  Later work improves OPD by changing teacher
contexts, objectives, or token weights
\cite{ye2026policy,yang2026learning,hou2026uni,jin2026entropy,xu2026tip}.
\opsd{} removes the need for a separate teacher by conditioning the same model
on privileged information, and related self-distillation methods use reference
solutions, environmental feedback, or successful rollouts as auxiliary guidance
\cite{zhao2026opsd,penaloza2026pid,shenfeld2026self,hubotter2026reinforcement}.
Recent extensions further calibrate, filter, route, relax, or localize this
supervision
\cite{nguyen2026avsd,zheng2026scope,yu2026dopd,tu2026ucob,ko2026reopold,shen2026purified,zhao2026rosd},
while analyses study scale, rollout quality, and optimization
\cite{li2026rethinkingopd,kaur2026rethinkingsd}.  DAPD retains on-policy
sampling and privileged-teacher construction, but reorganizes supervision
around matched information availability and both completion sources.

\section{Conclusion}

We identify information asymmetry as the root cause of privilege illusion in
\opsd{} and propose DAPD to address it through two forms of anchoring.  DPA
aligns reference and rollout behavior under matched information availability,
while DSA combines correctness-oriented reference guidance with
student-reachable rollout guidance.  Across benchmarks and Qwen3 models from 1.7B to 32B, DAPD significantly
alleviates privilege illusion and consistently outperforms \opsd{}, validating
the benefits of both matched-information paths and guidance sources.

\paragraph{Limitations.}
DAPD constructs multiple anchored distributions during training, increasing
training-time computation but adding no inference-time cost.  Its weights may
need recalibration across scales or architectures.  The verified-rollout
extension requires an automatic correctness signal, which may be unavailable for
open-ended tasks.

\clearpage
\appendix
\section{Experimental Details}
\label{app:experimental-details}

This section specifies the data, evaluation, privileged-information
construction, training procedure, baseline implementations, and final
DAPD configurations used in the paper.

\subsection{Benchmarks and Evaluation}

For AIME 2024, AIME 2025 \cite{maa2026aime}, and HMMT February 2025
\cite{hmmt2025february}, we sample 12
thinking-enabled solutions per problem at temperature 1.0 and top-$p$ 0.95,
allow up to 38,912 new tokens, and verify the final answer.  Avg@12 averages
correctness over the 12 samples for each problem.  The reported reasoning
aggregate is the unweighted mean of the three benchmark-level scores.

For broader evaluation, LiveCodeBench v5 \cite{jain2025livecodebench} uses its
official code-generation split and Pass@1 evaluator, BFCL v3
\cite{patil2025bfcl} uses the official multi-turn function-call harness and
overall accuracy, and IFBench \cite{pyatkin2025ifbench} uses the released
verifier-based IFBench score.  We retain each benchmark's official prompt and
parsing rules. Each evaluation uses the model trained under its
capability-specific data protocol below.

\subsection{Training Data}

All task-specific models use OpenThoughts data.  The Reasoning protocol uses
the math domain, the Coding protocol uses the code domain, and the Instruct
protocol uses the released mixture of math, code, and science
domains \cite{guha2025openthoughts}.  Within each protocol, every baseline uses
the same data split and prompt format.  Benchmark evaluation examples are
excluded from every training corpus.

\subsection{Construction of Privileged Information}
\label{app:privileged-construction}

We follow the reference-injection format of \opsd{}
\cite{zhao2026opsd}.  For each problem $x$, the reference completion $y^*$
is inserted into the teacher's user message between explicit delimiters.
The student rollout prompt contains only $x$.  Table~\ref{tab:view-templates}
shows the resulting user messages.  The same chat template is then applied to
both.

\begin{table}[!ht]
\centering
\footnotesize
\setlength{\tabcolsep}{4pt}
\begin{tabular}{@{}>{\raggedright\arraybackslash}p{0.13\columnwidth}
                >{\raggedright\arraybackslash}p{0.79\columnwidth}@{}}
\toprule
View & User-message template \\
\midrule
\textbf{\None{}}
&
\textbf{Problem:} \texttt{\{problem\}}

\smallskip
Please reason step by step, and put your final answer within
\texttt{\textbackslash boxed\{\}}.
\\
\addlinespace[4pt]
\textbf{\Cross{}}
&
\textbf{Problem:} \texttt{\{problem\}}

\smallskip
Here is a reference solution to this problem:

\texttt{=== Reference Solution Begin ===}

\texttt{\{reference completion\}}

\texttt{=== Reference Solution End ===}

\smallskip
After reading the reference solution above, make sure you truly understand
the reasoning behind each step --- do not copy or paraphrase it.  Now, using
your own words and independent reasoning, derive the same final answer to the
problem above.  Think step by step, explore different approaches, and don't
be afraid to backtrack or reconsider if something doesn't work out:

\smallskip
Please reason step by step, and put your final answer within
\texttt{\textbackslash boxed\{\}}.
\\
\bottomrule
\end{tabular}
\caption{User-message templates for the \None{} view without privileged information and the
reference-conditioned \Cross{} view.  Braced fields are replaced per example.}
\label{tab:view-templates}
\end{table}

For either completion $s\in\{y,y^*\}$, DAPD computes token-level supervision
on the tokens of $s$.  \Cross{} inserts the other completion $\bar{s}$ into the
privileged message, while \Self{} inserts $s$ itself.  Both distributions are
evaluated on the prefixes and target tokens of $s$.  These auxiliary
completions are available only when constructing training distributions and
are never included in the inference prompt.

\paragraph{Concrete training example.}
One OpenThoughts example asks to order
$p=2^{3009}$, $q=3^{2006}$, and $r=5^{1003}$.  Its reference completion
rewrites them as $8^{1003}$, $9^{1003}$, and $5^{1003}$, respectively, and
concludes $r<p<q$.  The \None{} message contains only the question and answer
instruction.  The \Cross{} and \Self{} messages insert this complete derivation
between the two reference-solution delimiters above.  Thus the privileged
information is an explicit worked completion, rather than a label or latent
teacher state.

\subsection{DAPD Training Procedure}
\label{app:training-procedure}

For each minibatch, DAPD performs the following operations:
\begin{enumerate}
    \item Sample an on-policy rollout $y$ from the current LoRA-on student
    \cite{hu2022lora} for each prompt--reference pair $(x,y^*)$.
    \item For each source completion $s\in\{y,y^*\}$, construct
    $p_{\mathrm{None}}^s$, $p_{\mathrm{Cross}}^s$, and
    $p_{\mathrm{Self}}^s$ on the tokens of $s$.
    \item Evaluate Entangled Distillation, Inference Anchor, and Privileged
    Anchor in both completion directions.  Detach every teacher distribution
    and retain gradients only through the student side.
    \item Combine the objective losses using the coefficients in
    Equation~\eqref{eq:main-objective-weights}, backpropagate their sum, and update
    the current LoRA parameters.
    \item Periodically copy the current LoRA parameters into the shared
    snapshot used by the two Entangled-Distillation teachers.
\end{enumerate}
The current LoRA-on policy produces every rollout and trainable distribution.
The two Entangled-Distillation teachers use the shared snapshot.  The
Inference-Anchor and Privileged-Anchor teachers use the detached LoRA-off base
policy.

\subsection{Baseline Methods}
\label{app:baseline-methods}

We compare methods that use the same student backbone and on-policy data
budget but differ in how they construct and transfer supervision.
\begin{itemize}
    \item \textbf{Base} is the original Qwen3 policy \cite{yang2025qwen3}
    evaluated with the task prompt.
    \item \textbf{\opsd{}} \cite{zhao2026opsd} samples from the current
    student and distills the same-model \Cross{} distribution into \None{} on the
    sampled prefixes.
    \item \textbf{SDFT} \cite{shenfeld2026self} uses on-policy samples and an
    exponential-moving-average self-teacher conditioned on an expert
    demonstration.
    \item \textbf{SDPO} \cite{hubotter2026reinforcement} conditions an
    exponential-moving-average self-teacher on successful peer rollouts or
    environment feedback to provide dense token-level supervision.
    \item \textbf{Purified \opsd{}} \cite{shen2026purified} removes the
    component of privileged supervision that can be predicted from the
    reference alone through a pointwise-mutual-information correction.
    \item \textbf{DOPD} \cite{yu2026dopd} dynamically routes token-level
    supervision between privileged teacher and student policies according to
    their advantage gap and relative probabilities.
\end{itemize}

\subsection{Training and Loss Implementation}

\paragraph{Optimization.}
We adapt all attention and
multilayer-perceptron projections with LoRA rank 64
and scale 128.  The optimizer uses a learning rate of $5\times10^{-6}$, a
linear 500-step schedule without warmup, gradient-norm clipping at 0.1,
bfloat16, gradient checkpointing, and an effective batch size of 32.

\paragraph{Random seeds.}
Unless otherwise specified, all training jobs use seed 42, set before model and
LoRA-adapter initialization, data shuffling, and rollout sampling.  Matched
method comparisons use the same seed so that initialization, data order, and
rollout-generation streams are aligned.  Reported headline numbers follow this
matched-seed protocol.

\paragraph{Compute and software.}
Training uses eight NVIDIA A100-80GB GPUs.  The software
environment uses PyTorch 2.8.0, Transformers 4.57.1, DeepSpeed 0.18.2, and
vLLM 0.11.0 \cite{kwon2023vllm}.  Data parallelism changes the per-device
microbatch as needed but keeps the effective batch size fixed.

The Coding and Instruct comparisons retain the same LoRA parameterization,
optimizer, update budget, and DAPD objective within each model scale.  Dataset
parsing and prompt formatting are adapted to the corresponding OpenThoughts
examples, and every baseline is rerun under the same capability-specific
protocol.

\paragraph{Rollout generation.}
The student samples one rollout per problem at temperature 1.1,
top-$p$ 0.95, top-$k$ 20, and a maximum of 1,024 new tokens.  The training
context is limited to 20,000 tokens.

\paragraph{Divergence implementation.}
Every directed loss term evaluates the full vocabulary at each non-padding
completion token.  Teacher and student logits are divided by $T=1.1$.  The
divergence $\Div(\sg[q]\|p)$ lists teacher $q$ before student $p$.  With
component cap $c=0.05$, we implement it as
\begin{equation}
\ell_c(q,p)=\sum_{v\in\mathcal V}
\min\{q_v(\log q_v-\log p_v),c\}.
\label{eq:clipped-kl}
\end{equation}
This is a component-clipped forward-KL surrogate.
Both distributions are evaluated directly at every sampled prefix, so no
importance-sampling ratio is applied.

\subsection{Main DAPD Configurations}
\label{app:main-configs}
\label{app:snapshot-strategy}

The main objective applies all three directed losses in both guidance
directions.  The first tuple below is the reference-guided
Rollout-to-Reference direction, and the second is the rollout-guided
Reference-to-Rollout direction.  We parameterize their coefficients as
\begin{align}
(w_{\mathrm{ent}}^{\mathrm{ref}},w_{\mathrm{infer}}^{\mathrm{ref}},w_{\mathrm{priv}}^{\mathrm{ref}})
  &=\frac{\kappa\lambda}{1+\beta_{\mathrm{infer}}^{\mathrm{ref}}+\beta_{\mathrm{priv}}^{\mathrm{ref}}}
    (1,\beta_{\mathrm{infer}}^{\mathrm{ref}},\beta_{\mathrm{priv}}^{\mathrm{ref}}),\notag\\
(w_{\mathrm{ent}}^{\mathrm{roll}},w_{\mathrm{infer}}^{\mathrm{roll}},w_{\mathrm{priv}}^{\mathrm{roll}})
  &=\frac{\kappa(1-\lambda)}{1+\beta_{\mathrm{infer}}^{\mathrm{roll}}+\beta_{\mathrm{priv}}^{\mathrm{roll}}}
    (1,\beta_{\mathrm{infer}}^{\mathrm{roll}},\beta_{\mathrm{priv}}^{\mathrm{roll}}).
\label{eq:main-objective-weights}
\end{align}
Here $\kappa$ controls the total DAPD loss weight.  The reference-guided
direction receives fraction $\lambda$, and the rollout-guided direction
receives $1-\lambda$.  The four $\beta$ values redistribute weight among the
three loss terms within each direction.  The Privileged Anchor ablation fixes
$\beta_{\mathrm{infer}}^{\mathrm{ref}}=\beta_{\mathrm{infer}}^{\mathrm{roll}}=1$
and varies $\beta_{\mathrm{priv}}^{\mathrm{ref}}$ and
$\beta_{\mathrm{priv}}^{\mathrm{roll}}$.  The Inference Anchor ablation fixes
$\beta_{\mathrm{priv}}^{\mathrm{ref}}=\beta_{\mathrm{priv}}^{\mathrm{roll}}=1$
and varies $\beta_{\mathrm{infer}}^{\mathrm{ref}}$ and
$\beta_{\mathrm{infer}}^{\mathrm{roll}}$.

Every teacher distribution is detached.  Only the two Entangled-Distillation
instances are produced by one shared snapshot of the LoRA-on student, which is
copied and periodically updated during training.  The teacher distributions of
the Inference Anchor and Privileged Anchor terms are produced by the LoRA-off
base model.  All student distributions are produced by the current LoRA-on
model, which also serves as the rollout sampler.

\begin{table}[!ht]
\centering
\small
\setlength{\tabcolsep}{3.3pt}
\begin{tabular}{@{}lccccc@{}}
\toprule
& & \multicolumn{2}{c}{Reference} & \multicolumn{2}{c}{Rollout} \\
\cmidrule(lr){3-4}\cmidrule(lr){5-6}
Scale & $\lambda$ & $\beta_{\mathrm{infer}}$ & $\beta_{\mathrm{priv}}$
& $\beta_{\mathrm{infer}}$ & $\beta_{\mathrm{priv}}$ \\
\midrule
1.7B & .5 & 1 & .5 & 1 & .5 \\
4B   & .2 & 1 & 1  & 1 & 2 \\
8B   & .2 & 1 & 2  & 1 & .5 \\
14B  & .2 & 1 & 2  & 1 & 1 \\
32B  & .2 & 1 & .5 & 1 & 1 \\
\bottomrule
\end{tabular}
\caption{Coefficient configurations for the scale-wise DAPD results.}
\label{tab:main-config-details}
\end{table}

The task-specific Qwen3-4B Reasoning, Coding, and Instruct models use the 4B
coefficient allocation above with their corresponding data protocol.

\section{Theoretical Analysis of the Unconditioned Path}
\label{app:path-analysis}

The Rollout-to-Reference unconditioned path ($y\to y^*$) combines
Inference Anchor on reference prefixes with Entangled Distillation on rollout
prefixes.  We formalize how \Self{} connects these two objectives and supports
the implicit alignment between the two \None{} distributions claimed in the
main paper.  The analysis has three parts: (i) a paired-prefix alignment bound,
(ii) sequence-level transfer on inference-policy states, and (iii) the
parameter-sharing mechanism that carries the Inference Anchor update across
prefixes.

\subsection{Conditional Alignment of the \None{} Distributions}

The unconditioned path aims to align the rollout- and reference-side \None{}
distributions, although its two objectives act through \Cross{} and \Self{}.
We first show why these intermediate distributions suffice: small objective
values and a consistent \Self{}--\Cross{} bridge bound the desired
\None{}--\None{} distance.  Fix a
reference prefix $a=y^*_{<i}$ and a rollout prefix $b=y_{<j}$, and abbreviate
the four token distributions as
\begin{align}
n^*&=p_\theta(\cdot\mid x,a),
&s^*&=p_\theta(\cdot\mid x,a,y^*),\notag\\
c^y&=p_\theta(\cdot\mid x,b,y^*),
&n^y&=p_\theta(\cdot\mid x,b).
\label{eq:paired-prefix-distributions}
\end{align}
These are, respectively, \None{} and \Self{} on the reference prefix and \Cross{} and
\None{} on the rollout prefix.  Let $\Pi_x$ be any coupling of the reference- and
rollout-prefix sampling distributions.  Define
\begin{align}
\mathcal E_{\mathrm{infer}}^\Pi&=
\mathbb E_{(a,b)\sim\Pi_x}\KL(n^*\|s^*),\notag\\
\mathcal E_{\mathrm{ent}}^\Pi&=
\mathbb E_{(a,b)\sim\Pi_x}\KL(c^y\|n^y),\notag\\
\epsilon_{\mathrm{bridge}}&=
\mathbb E_{(a,b)\sim\Pi_x}\operatorname{TV}(s^*,c^y).
\label{eq:bridge-quantities}
\end{align}
Here $\operatorname{TV}$ denotes total variation distance.

\paragraph{Bridge-consistency assumption.}
For coupled reference and rollout prefixes from the same example, assume
\begin{equation}
\epsilon_{\mathrm{bridge}}\leq \epsilon.
\label{eq:bridge-consistency}
\end{equation}
This assumption captures the proxy relation used by DPA: \Self{} and \Cross{}
are produced by the same policy and share the privileged condition $(x,y^*)$.

\paragraph{Proposition 1 (paired-prefix \None{}-distribution alignment).}
The two \None{} distributions satisfy
\begin{equation}
\mathbb E_{\Pi_x}\operatorname{TV}(n^*,n^y)
\leq \sqrt{\frac{\mathcal E_{\mathrm{infer}}^\Pi}{2}}+\epsilon
+\sqrt{\frac{\mathcal E_{\mathrm{ent}}^\Pi}{2}}.
\label{eq:none-distribution-bound}
\end{equation}

\paragraph{Proof.}
For every paired prefix, the triangle inequality gives
\begin{equation*}
\operatorname{TV}(n^*,n^y)
\leq \operatorname{TV}(n^*,s^*)
+\operatorname{TV}(s^*,c^y)
+\operatorname{TV}(c^y,n^y).
\end{equation*}
Apply Pinsker's inequality to the first and third terms, average under
$\Pi_x$, apply Jensen's inequality to each square root, and use
Eq.~\eqref{eq:bridge-consistency}.

Proposition~1 directly connects the two terms in the unconditioned path to its
target alignment.  Inference Anchor reduces the first term, Entangled
Distillation reduces the third, and the shared \Self{}--\Cross{} bridge controls
the middle term.  Their joint optimization therefore tightens an explicit
upper bound on the distance between
$p_{\mathrm{None}}^{y^*}$ and $p_{\mathrm{None}}^y$.

\subsection{Sequence-Level Transfer from Entangled Distillation}

The preceding result establishes token-level conditional alignment, but it
does not yet show that complete inference trajectories inherit \Cross{}
behavior.  We therefore lift Entangled Distillation from token conditionals
on on-policy prefixes to a bound on complete sequence distributions.  Fix an
input $x$, a guiding completion $r$, and a horizon $H$, appending an
absorbing end-of-sequence token to shorter completions.  Let
$N_\theta(\cdot\mid x)$ be the sequence distribution induced by \None{} and let
$P_\theta^r(\cdot\mid x)$ be the sequence distribution induced by \Cross{} when
conditioned on $r$.  For a rollout $Y\sim N_\theta(\cdot\mid x)$, define the
ideal summed-token Entangled Distillation loss, suppressing fixed $x$ and
$\theta$ and writing the token conditionals at $Y_{<t}$ as $P_t^r$ and $N_t$,
\begin{equation}
\mathcal E_{\mathrm{ent}}(x,r)
=\mathbb E_Y\!\left[\sum_{t=1}^{H}\KL(P_t^r\|N_t)\right].
\label{eq:appendix-ent-loss}
\end{equation}

\paragraph{Lemma 1 (autoregressive hybrid bound).}
For any two length-$H$ autoregressive distributions $U$ and $V$, write their
next-token conditionals at $Z_{<t}\sim U$ as $U_t$ and $V_t$.  Then
\begin{equation}
\operatorname{TV}(U,V)
\leq\sum_{t=1}^{H}\mathbb E_{Z_{<t}\sim U}\!\left[\operatorname{TV}(U_t,V_t)\right].
\label{eq:autoregressive-hybrid}
\end{equation}

\paragraph{Proof.}
For $t=0,\ldots,H$, define
\begin{equation}
H^{(t)}(z_{1:H})
=\prod_{k=1}^{t}U_k(z_k\mid z_{<k})
 \prod_{k=t+1}^{H}V_k(z_k\mid z_{<k}).
\end{equation}
Thus $H^{(0)}=V$ and $H^{(H)}=U$.  The pair
$H^{(t-1)},H^{(t)}$ has the same $U$-induced distribution over $z_{<t}$,
uses $V_t$ or $U_t$ at position $t$, respectively, and then applies the same
$V$ continuation kernel.  Total variation contracts under this common
kernel, giving
\begin{equation*}
\operatorname{TV}\!\left(H^{(t-1)},H^{(t)}\right)
\leq
\mathbb E_{Z_{<t}\sim U}
\left[\operatorname{TV}(V_t,U_t)\right].
\end{equation*}
Summing consecutive hybrid distances proves
Eq.~\eqref{eq:autoregressive-hybrid}.

\paragraph{Proposition 2 (on-policy sequence transfer).}
The sequence distributions induced by \None{} and \Cross{} satisfy
\begin{equation}
\operatorname{TV}\!\left(
N_\theta(\cdot\mid x),P_\theta^r(\cdot\mid x)
\right)
\leq
\sqrt{\frac{H}{2}\mathcal E_{\mathrm{ent}}(x,r)}.
\label{eq:ent-sequence-transfer}
\end{equation}

\paragraph{Proof.}
Apply Lemma~1 with $U=N_\theta$ and $V=P_\theta^r$.  At each prefix sampled
from $N_\theta$, Pinsker's inequality gives
\begin{equation*}
\operatorname{TV}\!\left(N_{\theta,t},P_{\theta,t}^r\right)
\leq
\sqrt{\tfrac12\KL(P_{\theta,t}^r\|N_{\theta,t})}.
\end{equation*}
The local KL is algebraically $\KL(\mathit{Cross}\|\mathit{None})$, while its
student-to-teacher alignment arrow is $\mathit{None}\to\mathit{Cross}$.
Total variation is symmetric, so Jensen's and Cauchy--Schwarz inequalities
convert the sum of local bounds into Eq.~\eqref{eq:ent-sequence-transfer}.

\subsection{Shared-Parameter Transfer from Inference Anchor}

The preceding result explains how Entangled Distillation transfers \Cross{}
behavior to rollout-side \None{}, but Inference Anchor is optimized on
reference-side \Self{} rather than rollout-side \Cross{}.  We therefore show
how shared parameters carry an Inference Anchor update from \Self{} to
\Cross{}, providing the optimization channel required by the proxy bridge.

At a reference prefix, let $q_{\mathrm N}$ be the detached \None{} teacher
distribution and
$p_{\mathrm S}$ the trainable \Self{} distribution.  At a rollout prefix, let
$p_{\mathrm C}$ be the \Cross{} distribution.  Write their pre-softmax logits as
$z_{\mathrm S}(\theta)$ and $z_{\mathrm C}(\theta)$, with Jacobians
$J_{\mathrm S}=\partial z_{\mathrm S}/\partial\theta$ and
$J_{\mathrm C}=\partial z_{\mathrm C}/\partial\theta$.  For the ideal
Inference Anchor
\begin{equation}
\ell_{\mathrm{infer}}=\KL(q_{\mathrm N}\|p_{\mathrm S}),
\end{equation}
the unit-temperature logit gradient is $p_{\mathrm S}-q_{\mathrm N}$.
Consequently, one gradient step with step size $\eta$ using only this loss term changes the \Cross{}
logits by
\begin{equation}
z_{\mathrm C}(\theta^+)-z_{\mathrm C}(\theta)
=-\eta J_{\mathrm C}J_{\mathrm S}^{\top}
(p_{\mathrm S}-q_{\mathrm N})+O(\eta^2).
\label{eq:cross-prefix-transfer}
\end{equation}

The cross-prefix kernel $J_{\mathrm C}J_{\mathrm S}^{\top}$ is the first-order
channel from grounding \Self{} on reference tokens to changing \Cross{} on
rollout tokens.  Let $g_{\mathrm S}=p_{\mathrm S}-q_{\mathrm N}$ and
$g_{\mathrm C}=p_{\mathrm C}-q_{\mathrm N}$.  Taking expectation over examples
and their paired-prefix coupling, we use the local compatibility condition
\begin{equation}
\mathbb E_{x,(a,b)\sim\Pi_x}\!\left[
g_{\mathrm C}^{\top}J_{\mathrm C}J_{\mathrm S}^{\top}g_{\mathrm S}
\right]
\geq \gamma>0.
\label{eq:cross-prefix-compatibility}
\end{equation}
Because $g_{\mathrm C}$ is the logit gradient of
$\operatorname{CE}(q_{\mathrm N},p_{\mathrm C})$, Eqs.~\eqref{eq:cross-prefix-transfer}
and \eqref{eq:cross-prefix-compatibility} give the expected change
\begin{equation}
\mathbb E\!\left[
\operatorname{CE}(q_{\mathrm N},p_{\mathrm C}(\theta^+))
-\operatorname{CE}(q_{\mathrm N},p_{\mathrm C}(\theta))
\right]\leq-\eta\gamma+O(\eta^2).
\label{eq:cross-prefix-descent}
\end{equation}
Thus, a compatible Inference Anchor step moves \Cross{} in expectation toward the
reference-side \None{} anchor.  Refreshing the \Cross{} teacher then carries
this anchored change into rollout-side \None{} through Entangled
Distillation.  Shared parameters provide the optimization channel underlying
the proxy bridge in Section~\ref{sec:dapd}.

\paragraph{Empirical compatibility check.}
We directly measure the scalar in
Eq.~\eqref{eq:cross-prefix-compatibility} in the trainable LoRA subspace of
Qwen3-4B.  Using four training-matched LoRA initializations and 32 distinct
math examples, we sample on-policy rollouts and compute full-vocabulary
gradients at four interior token positions per example.  Across the resulting
128 positions, the mean gradient dot product is $256.62$ with a bootstrap
$95\%$ confidence interval of $[147.55,393.53]$, and the mean gradient cosine
is $0.101$ with an interval of $[0.058,0.144]$.  After averaging the four
positions within each example, the corresponding values are $225.30$
$[122.94,352.53]$ and $0.119$ $[0.054,0.185]$.  The positive intervals across
both aggregation levels empirically support the expected local compatibility
used above.

\subsection{Implication for DAPD}

The three results support the unconditioned-path construction from
complementary perspectives.  Proposition~1 bounds the desired
\None{}-to-\None{} alignment through the two trained objectives and the
\Self{}--\Cross{} bridge.  Proposition~2 shows that Entangled Distillation
transfers the resulting teacher behavior on on-policy prefixes, while
Eq.~\eqref{eq:cross-prefix-descent} explains how a compatible Inference Anchor
step shapes that teacher through shared parameters.  Together, they formalize
why the joint update implicitly moves $p_{\mathrm{None}}^y$ toward
$p_{\mathrm{None}}^{y^*}$ under the stated bridge-consistency and local
compatibility conditions.

\section{Additional Experimental Results}
\label{app:additional-results}

This section provides the absolute scale-wise scores, the measurements behind
the privilege-illusion analysis, and implementation sensitivity.

\subsection{Complete Scale-Wise Results}

Table~\ref{tab:scale-results-full} expands Figure~\ref{fig:teaser}(b) into
per-benchmark scores.  Every DAPD row uses the complete objective and the
configuration in Table~\ref{tab:main-config-details}.

\begin{table*}[!ht]
\centering
\small
\setlength{\tabcolsep}{8pt}
\begin{tabular}{@{}llrrrr@{}}
\toprule
Scale & Method & AIME24 & AIME25 & HMMT25 & Avg@12 \\
\midrule
\multirow{3}{*}{1.7B}
& Base   & 50.83 & 37.78 & 21.94 & 36.85 \\
& \opsd{} & 57.78 & 40.56 & 27.78 & 42.04 \\
& DAPD   & 57.50 & 46.67 & 27.78 & \textbf{43.98} \\
\midrule
\multirow{3}{*}{4B}
& Base   & 75.56 & 65.56 & 42.50 & 61.20 \\
& \opsd{} & 76.67 & 67.78 & 43.33 & 62.59 \\
& DAPD   & 77.22 & 72.22 & 46.39 & \textbf{65.28} \\
\midrule
\multirow{3}{*}{8B}
& Base   & 80.56 & 67.78 & 46.67 & 65.00 \\
& \opsd{} & 80.56 & 67.50 & 46.94 & 65.00 \\
& DAPD   & 80.00 & 72.50 & 49.72 & \textbf{67.41} \\
\midrule
\multirow{3}{*}{14B}
& Base   & 84.17 & 70.83 & 51.39 & 68.80 \\
& \opsd{} & 81.39 & 72.78 & 52.50 & 68.89 \\
& DAPD   & 85.83 & 75.56 & 51.39 & \textbf{70.93} \\
\midrule
\multirow{3}{*}{32B}
& Base   & 83.61 & 73.06 & 53.33 & 70.00 \\
& \opsd{} & 85.28 & 75.28 & 50.28 & 70.28 \\
& DAPD   & 86.11 & 76.67 & 56.39 & \textbf{73.06} \\
\bottomrule
\end{tabular}
\caption{Complete reasoning results underlying the scale comparison.  Avg@12
is the unweighted mean of AIME24, AIME25, and HMMT25.}
\label{tab:scale-results-full}
\label{tab:scale-results}
\end{table*}

\begin{table}[!ht]
\centering
\small
\setlength{\tabcolsep}{4.5pt}
\begin{tabular}{@{}rccc@{}}
\toprule
Step & \opsd{} & Privileged Anchor & \dapd{} \\
\midrule
\multicolumn{4}{@{}l}{\textit{Wrong claims per 10k}} \\
0   & 7.41  & 7.41  & 9.26 \\
50  & 7.41  & 16.67 & 9.26 \\
100 & 14.81 & 3.70  & 1.85 \\
150 & 11.11 & 7.41  & 3.70 \\
200 & 12.96 & 14.81 & 7.41 \\
250 & 24.07 & 11.11 & 5.56 \\
300 & 37.04 & 22.22 & 11.11 \\
\midrule
\multicolumn{4}{@{}l}{\textit{Reasoning Avg@12}} \\
0   & 60.56 & 60.56 & 60.56 \\
50  & 60.83 & 62.04 & 63.65 \\
100 & 61.50 & 61.82 & 63.11 \\
150 & 60.59 & 61.15 & 62.70 \\
200 & 59.56 & 61.67 & 63.10 \\
250 & 56.96 & 59.93 & 61.36 \\
300 & 53.24 & 59.46 & 60.90 \\
\bottomrule
\end{tabular}
\caption{Five-scale training dynamics underlying
Figure~\ref{fig:privilege-dynamics}.  The Privileged Anchor replaces the
trainable \None{} distribution with \Self{} while retaining the \Cross{}
teacher, and DAPD combines both anchoring paths and guidance sources.}
\label{tab:privilege-dynamics-values}
\end{table}

\begin{table}[!ht]
\centering
\small
\setlength{\tabcolsep}{7pt}
\begin{tabular}{@{}lcc@{}}
\toprule
Divergence & Component cap & Avg@12 \\
\midrule
Forward KL & 0.05 & \textbf{65.28} \\
Forward KL & None & 62.50 \\
Reverse KL & 0.05 & 62.50 \\
Reverse KL & None & 63.33 \\
\bottomrule
\end{tabular}
\caption{Divergence and component-clipping ablation on Qwen3-4B.}
\label{tab:divergence-clipping}
\end{table}

\begin{table}[!t]
\centering
\scriptsize
\setlength{\tabcolsep}{3pt}
\renewcommand{\arraystretch}{1.05}
\begin{tabular}{@{}p{0.18\columnwidth}p{0.76\columnwidth}@{}}
\toprule
Policy & Output excerpt \\
\midrule
\multicolumn{2}{@{}l}{\textbf{Qwen3-4B on AIME25 \#20}} \\
\opsd{} ($6/12$)
& After leaving the arc relations unresolved, the output states:
\emph{``since the problem is from a competition, the answer is likely to be
a nice number, like $360^\circ$''}, and returns $360$. \\
\dapd{} ($12/12$)
& The output obtains $\widehat{DE}=72^\circ$, $\widehat{HJ}=24^\circ$, and
$\widehat{FG}=72^\circ$, then computes
$72+2(24)+3(72)=\boxed{336}$. \\
\addlinespace[2pt]
\multicolumn{2}{@{}l}{\textbf{Qwen3-4B on HMMT25 \#27}} \\
\opsd{} ($8/12$)
& After finding that its candidate violates the circle equation, the output
states: \emph{``I recall that the answer is likely $1/7$ or $2/7$''} and
returns $\boxed{1/7}$. \\
\dapd{} ($11/12$)
& The output derives $A=(3/2,3\sqrt{15}/2)$ and
$P=(7/4,7\sqrt{15}/12)$. It computes
$[BPC]=49\sqrt{15}/24$ and $[ABC]=21\sqrt{15}/4$, giving
$\boxed{7/18}$. \\
\addlinespace[2pt]
\multicolumn{2}{@{}l}{\textbf{Qwen3-8B on AIME25 \#9}} \\
\opsd{} ($7/12$)
& After obtaining a sign incompatible with the required answer form, the
output states: \emph{``the answer is likely to be $50$, I will go with this,
assuming there was a sign error in the calculation''}, and returns $50$. \\
\dapd{} ($10/12$)
& The output factors $t^4+2t^3-25t^2-18t+120$ as
$(t^2-t-10)(t^2+3t-12)$, derives $y=(3-\sqrt{57})/2$, and returns
$3+57+2=\boxed{62}$. \\
\bottomrule
\end{tabular}
\caption{Additional inference-time cases.  Counts report correct outputs
among 12 generations.  Quoted clauses are verbatim, while the remaining text
summarizes omitted reasoning.}
\label{tab:additional-cases}
\end{table}

DAPD improves over the scale-matched \opsd{} policy by 1.94, 2.69, 2.41,
2.04, and 2.78 points from 1.7B through 32B, respectively.  The gains are not
concentrated in one benchmark: DAPD improves AIME25 at every scale and
typically improves HMMT25 as well.

\subsection{Privilege-Illusion Dynamics}
\label{app:privilege-dynamics}

\paragraph{Behavioral probe.}
The primary probe detects a generation that first states that its derivation
has failed and then attributes a concrete answer to memory, a reference or
official solution, or an external source, while returning an incorrect final
answer.  Generic uncertainty, ordinary guesses, recalled formulas, and broad
phrases such as ``after research'' are excluded.

Each scale--checkpoint cell contains 90 benchmark problems and 12 generations
per problem.  We report counts per 10,000 generations after pooling the five
Qwen3 scales.  Confidence intervals use a problem-cluster bootstrap that keeps
all 12 generations of a problem together.

Across all checkpoints, \opsd{} produces 58 detected claims in 32,400
generations, compared with 41 for the Privileged Anchor.  At steps 250--300,
the counts are 33 and 18, a 45\% reduction.

At steps 250--300, DAPD produces 9 detected claims compared with 33 for
\opsd{}, a 73\% reduction.  Its step-300 Avg@12 is also 7.66 points higher.
Together with the isolated Privileged-Anchor intervention, these dynamics show
that matched-information anchoring reduces privilege illusion while preserving
late-stage reasoning performance.

\subsection{Divergence and Clipping}
\label{app:divergence-clipping}

To examine sensitivity to divergence direction and clipping, we summarize
completed Qwen3-4B results for the evaluated implementations.

Component-clipped forward KL performs best in this comparison, supporting its
use as the default DAPD implementation.

\section{Additional Qualitative Evidence}
\label{app:additional-cases}

Table~\ref{tab:additional-cases} supplements
Figure~\ref{fig:full-method-case} with three inference-time comparisons.
Neither method receives a reference at inference, and both use the same
reference-free prompt and decoding protocol.  The counts report
correctness over all 12 samples rather than only the displayed output.
\FloatBarrier

\section{Future Directions}
\label{app:future-directions}

\paragraph{Reference-free DAPD.}
The dual-rollout experiments in Section~\ref{sec:experiments} show that a
second rollout can provide useful guidance even without a reference
completion.  A natural next step is to replace the fixed reference with a
verified or consensus-selected rollout and update this source as the policy
improves.  This would preserve the matched-information paths while reducing
dependence on curated solutions.

\paragraph{Adaptive source trust.}
The scale-wise results indicate that reference and rollout guidance should
not receive a universal fixed balance.  Larger policies generate more useful
rollouts, but reliability also varies across examples and tokens.  Future
work can estimate this reliability online and assign continuous source and
anchor weights, avoiding both a global coefficient and a brittle hard gate.

\paragraph{Broader forms of privileged information.}
References are only one source of training-time information.  The same
framework can be studied with retrieved documents, tool traces, verifier
feedback, or intermediate plans that are unavailable at inference.  Studying
these sources can clarify how anchor design should adapt to the form and
reliability of privileged information.

\clearpage
\bibliography{references}

\end{document}